\pdfoutput=1
\documentclass[11pt]{article}

\usepackage[]{acl}

\usepackage{times}
\usepackage{latexsym}
\usepackage[T1]{fontenc}
\usepackage[utf8]{inputenc}
\usepackage{microtype}

\usepackage{xcolor}
\usepackage{hyperref}
\usepackage{graphicx}
\usepackage{adjustbox}
\usepackage{amssymb}
\usepackage{pifont}
\usepackage{multirow}
\usepackage{amsmath}
\usepackage{bbm}
\usepackage{float}
\usepackage{caption}
\usepackage{subcaption}

\title{
Just Rank: Rethinking Evaluation with Word and Sentence Similarities
}

\author{
Bin Wang\textsuperscript {\rm \dag}, 
C.-C. Jay Kuo\textsuperscript{\rm \S}, 
Haizhou Li\textsuperscript{\rm \ddag,\rm \dag, $\sharp$}
\\
\textsuperscript{\rm \dag} National University of Singapore, Singapore
\\
\textsuperscript{\rm \S} University of Southern California, USA
\\
\textsuperscript{\rm \ddag } The Chinese University of Hong Kong, Shenzhen, China
\quad
\textsuperscript{\rm $\sharp$} Kriston AI, China
\\
\texttt{bwang28c@gmail.com}} 

\begin{document}
\maketitle
\begin{abstract}
Word and sentence embeddings are useful feature representations in natural language processing.
However, intrinsic evaluation for embeddings lags far behind, and there has been no significant update since the past decade. 
Word and sentence similarity tasks have become the \textit{de facto} evaluation method.
It leads models to overfit to such evaluations, 
 negatively impacting embedding models' development.
This paper first points out the problems using semantic similarity as the gold standard for word and sentence embedding evaluations.
Further, we propose a new intrinsic evaluation method called \emph{EvalRank}, which shows a much stronger correlation with downstream tasks. 
Extensive experiments are conducted based on 60+ models and popular datasets to certify our judgments.  
Finally, the practical evaluation toolkit is released 
for future benchmarking purposes.\footnote{Available at \url{https://github.com/BinWang28/EvalRank-Embedding-Evaluation}.}
\end{abstract}

\section{Introduction}

    Distributed representation of words \cite{bengio2003neural,mikolov2013efficient,pennington2014glove,bojanowski2017enriching} and sentences \cite{kiros2015skip,conneau2017supervised,reimers2019sentence,gao2021simcse} have shown to be extremely useful in transfer learning to many NLP tasks. Therefore, it plays an essential role in how we evaluate the quality of embedding models. Among many evaluation methods, the word and sentence similarity task gradually becomes the de facto intrinsic evaluation method.
    
\begin{figure}[t]
    \centering
        \includegraphics[width=0.5\textwidth]{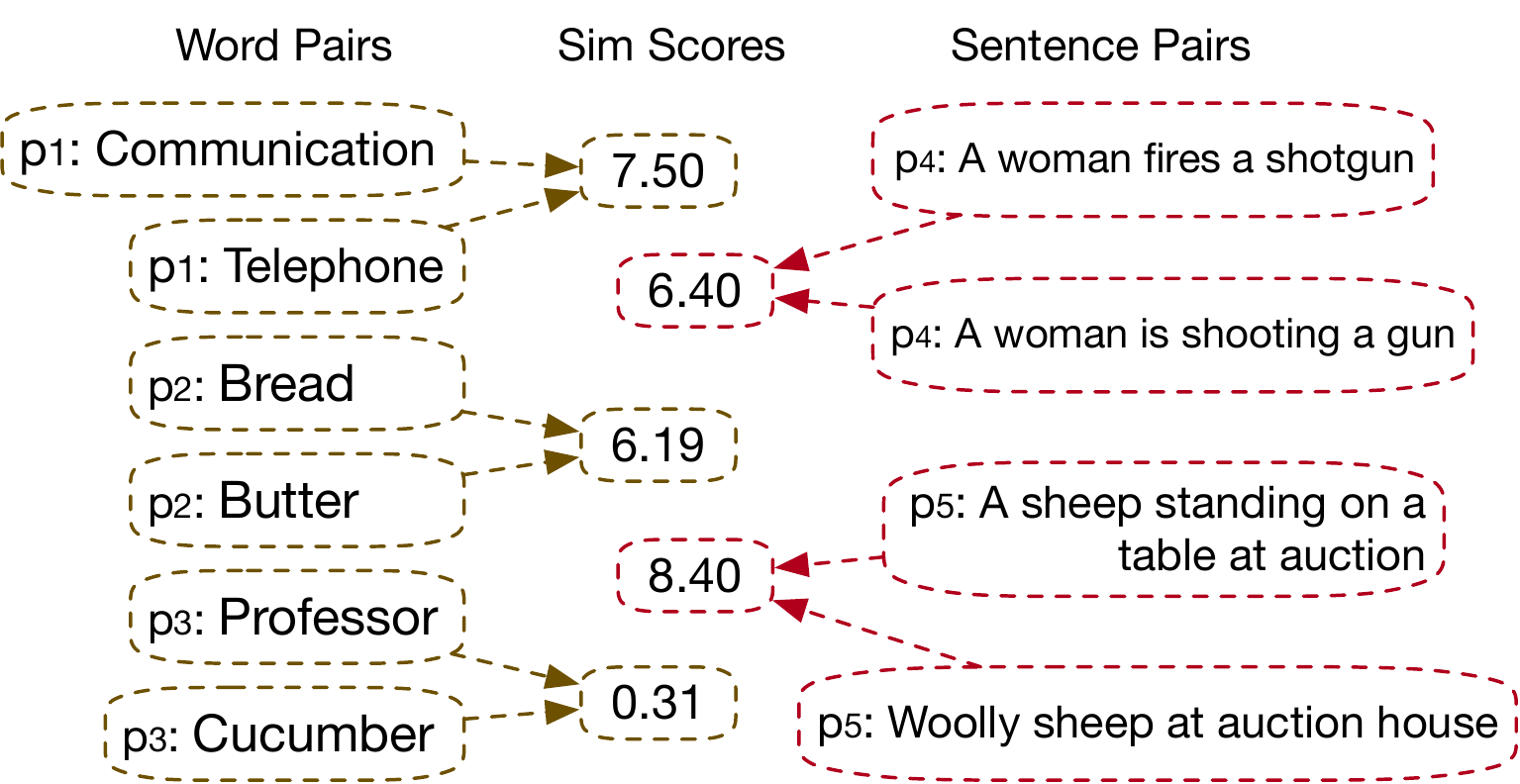}
    \caption{Word and sentence pairs with human-annotated similarity scores from WS-353 and STS-B datasets (scaled to range 0 (lowest) to 10 (highest)).}
    \label{fig:wspairs}
    \end{figure}

    Figure \ref{fig:wspairs} shows examples from word and sentence similarity datasets. 
    In general, the datasets consist of pairs of words ($w_1, w_2$) (or sentences) and human-annotated similarity scores $S_h$. 
    To evaluate an embedding model $\phi(\cdot)$, we first extract embeddings for ($w_1$,$w_2$):
    ($\mathbf{e}_{1}, \mathbf{e}_{2}$) = ($\phi(w_1),\phi(w_2)$).
    Then, a similarity measure is applied to compute an predicted score $S_p = sim(\mathbf{e}_{1}, \mathbf{e}_{2})$, where cosine similarity is adopted as $sim$ unquestionably in the majority of cases. Finally, the correlation between $S_h$ and $S_p$ is computed, and a higher correlation suggests good alignment with human annotations and a better embedding model.

    Many studies, especially those targeting on information retrieval via semantic search and clustering \cite{reimers2019sentence,su2021whitening}, have used the similarity task as the only or main evaluate method \cite{tissier2017dict2vec,mu2017all,arora2017simple,li2020sentence,gao2021simcse}. We observe a number of issues in word or sentence similarity tasks ranging from dataset collection to the evaluation paradigm, and consider that focusing too much on similarity tasks would negatively impact the development of future embedding models.
    
    The significant concerns are summarized as follows, which generally apply to both word and sentence similarity tasks.
    First, the definition of similarity is too vague.
    There exist complicated relationships between sampled data pairs, and almost all relations contribute to the similarity score, which is challenging to non-expert annotators.
    Second, the similarity evaluation tasks are not directly relevant to the downstream tasks.
    We believe it is because of the data discrepancy between them, and the properties evaluated by similarity tasks are not the ones important to downstream applications. 
    Third, the evaluation paradigm can be tricked with simple post-processing methods, making it unfair to benchmark different models.
    
    Inspired by Spreading-Activation Theory \cite{collins1975spreading}, we propose to evaluate embedding models as a retrieval task, and name it as \emph{EvalRank} to address the above issues.
    While similarity tasks measure the distance between similarity pairs from all similarity levels, \emph{EvalRank} only considers highly similar pairs from a local perspective.
    
    Our main contributions can be summarized as follows:
    \begin{itemize}
    
        \item[1] We point out three significant problems for using word and sentence similarity tasks as the de facto evaluation method through analysis or experimental verification. The study provides valuable insights into embeddings evaluation methods.
        \item[2] We propose a new intrinsic evaluation method,  \emph{EvalRank}, that aligns better with the properties required by various downstream tasks. 
        \item[3] We conduct extensive experiments with 60+ models and 10 downstream tasks to certify the effectiveness of our evaluation method. The practical evaluation toolkit is released for future benchmarking purposes.
        
    \end{itemize}

\section{Related Work}
    
    Word embedding has been studied extensively, and popular work \cite{mikolov2013efficient,pennington2014glove,bojanowski2017enriching} are mainly built on the distributional hypothesis \cite{harris1954distributional}, where words that appear in the same context tend to share similar meanings. The early work on sentence embedding are either built upon word embedding \cite{arora2017simple,ruckle2018concatenated,almarwani2019efficient} or follow the distributional hypothesis on a sentence level \cite{kiros2015skip,hill2016learning,logeswaran2018efficient}. Recent development of sentence embedding are incorporating quite different techniques including multi-task learning \cite{cer2018universal}, supervised inference data \cite{conneau2017supervised,reimers2019sentence}, contrastive learning \cite{zhang2020unsupervised,carlsson2020semantic,yan2021consert,gao2021simcse} and pre-trained language models \cite{li2020sentence, wang2020sbert, su2021whitening}. Nonetheless, even though different methods choose different evaluation tasks, similarity task is usually the shared task for benchmarking purposes.
    
    Similarity task is originally proposed to mimic human's perception about the similarity level between word or sentence pairs. The first word similarity dataset was collected in 1965 \cite{rubenstein1965contextual}, which consists of 65 word pairs with human annotations. It has been a standard evaluation paradigm to use cosine similarity between vectors for computing the correlation with human judges~\cite{agirre2009study}. Many studies raise concerns about such evaluation paradigm. \citet{faruqui2016problems} and \citet{wang2019evaluating} points out some problems with word similarity tasks, including low correlation with downstream tasks and lack of task-specific similarity. \citet{reimers2016task}, \citet{eger2019pitfalls} and \citet{zhelezniak2019correlation} states current evaluation paradigm for Semantic Textual Similarity (STS) tasks are not ideal. One most recent work \cite{abdalla2021makes} questions about the data collection process of STS datasets and creates a new semantic relatedness dataset (STR) by comparative annotations \cite{louviere1991best}. 

    There are also other intrinsic evaluation methods for word and sentence embedding evaluation, but eventually did not gain much popularity. Word analogy task is first proposed in \cite{mikolov2013efficient,mikolov2013linguistic} to detect linguistic relations between pairs of word vectors. \citet{zhu2020sentence} recently expanded the analogy concept to sentence level. However, the analogy task is more heuristic and fragile as an evaluation method \cite{gladkova2016analogy,rogers2017too}. Recently, probing tasks have been proposed to measure intriguing properties of sentence embedding models without worrying much about practical applications \cite{zhu2018exploring,conneau2018you,baranvcikova2019search}. Because of the lack of effective intrinsic evaluation methods, \citet{reimers2019sentence} and \citet{wang2021tsdae} seeks to include more domain-specific tasks for evaluation. 

\section{Problems with Similarity Tasks}
\label{sec:problems}

    In this work, we discuss the problems of similarity tasks both on word and sentence levels. They are highly similar from data collection to evaluation paradigm and are troubled by the same problems.
    
    \subsection{Multifaceted Relationships}
        
        First, the concept of similarity and relatedness are not well-defined. Similar pairs are related but not vise versa. Taking synonym, hypernym, and antonym relations as examples, the similarity rank should be ``synonym $>$ hypernym $>$ antonym'' while the relatedness rank should be ``synonym $>$ hypernym $\approx$ antonym''. This was not taken into consideration when constructing datasets. \citet{agirre2009study} intentionally split one word similarity dataset into similarity and relatedness subsets. However, we find that obtained subsets are erroneous towards polysemy, and the relatedness between pair (`stock', `egg', 1.81) is much lower than pair (`stock', `oil', 6.34).
        It is because only the `financial stock market' is compared but not the `stock of supermarkets`. 
        Furthermore, relationships between samples are far more complicated than currently considered, which is a challenge to all current datasets.
        
        Second, the annotation process is not intuitive to humans. The initial goal of the similarity task is to let the model mimic human perception. However, we found that the instructions on similarity levels are not well defined. For example, on STS 13$\sim$16 datasets, annotators must label sentences that `share some details' with a score of 2 and `on the same topic' with a score of 1. According to priming effect theory, \cite{meyer1971facilitation,weingarten2016primed}, humans are more familiar with ranking several candidate samples based on one pivot sample (priming stimulus). Therefore, a more ideal way of annotation is to give one pivot sample (e.g. `cup') and rank candidates with different similarity levels (e.g. `trophy', `tableware', `food', `article', `cucumber'). In other words, it is more intuitive for human to compare (a,b) $>$ (a,c) than (a,b) $>$ (c,d) as far as similarity is concerned. However, in practice, it is hard to collect a set of candidates for each pivot sample, especially for sentences.
        
    \begin{table}[t]
        \centering
        \begin{subtable}[b]{0.5\textwidth}
            \centering
            \begin{adjustbox}{width=1.0\textwidth,center}
                \begin{tabular}{  l | c | c | c }
                    \hline
                    Score (rank)    & STS-B     & SST2                                      & MR \\
                    \hline
                    GloVe           & 47.95 (4) & 79.52 (6\textcolor{red}{$\downarrow$})    & 77.54 (5\textcolor{red}{$\downarrow$})   \\
                    InferSent       & 70.94 (3) & 83.91 (3)                                 & 77.61 (4\textcolor{red}{$\downarrow$}) \\
                    BERT-cls        & 20.29 (6) & 86.99 (1\textcolor{green}{$\uparrow$})    & 80.99 (1\textcolor{green}{$\uparrow$}) \\
                    BERT-avg        & 47.29 (5) & 85.17 (2\textcolor{green}{$\uparrow$})    & 80.05 (2\textcolor{green}{$\uparrow$})   \\
                    BERT-flow       & 71.76 (2) & 80.67 (4\textcolor{red}{$\downarrow$})    & 77.01 (6\textcolor{red}{$\downarrow$})  \\
                    BERT-whitening  & 71.79 (1) & 80.23 (5\textcolor{red}{$\downarrow$})    & 77.96 (3\textcolor{red}{$\downarrow$}) \\
                    \hline
                    \end{tabular}
                \end{adjustbox}
                \caption{}
            \label{tab:overfitting-b}
        \end{subtable}
        \\
        \centering
        \begin{subtable}[b]{0.32\textwidth}
            \centering
            \begin{adjustbox}{width=1.0\textwidth,center}
            \begin{tabular}{  l | c | c  }
                \hline
                Rank & cos & $l_2$ \\
                \hline
                SBERT & 1 & 2\textcolor{red}{$\downarrow$} \\
                SimCSE & 2 & 1\textcolor{green}{$\uparrow$} \\
                BERT-avg & 5 & 3\textcolor{green}{$\uparrow$} \\
                BERT-flow & 4 & 4 \\
                BERT-whitening & 3 & 5\textcolor{red}{$\downarrow$}  \\
                \hline
                \end{tabular}
            \end{adjustbox}
            \caption{}
            \label{tab:overfitting-a}
        \end{subtable}
        \caption{(a) Performance scores and rank of embedding models on STS-B, SST2, and MR tasks. (b) Performance rank of models on STS-B testset with $cos$ and $l_2$ similarity metrics.}
        \label{fig:three graphs}
    \end{table}

    \subsection{Weak Correlation with Downstream Tasks}
    \label{pro:weak_corr}
        
        In previous studies, it was found that the performance of similarity tasks shows little or negative correlation with the performance of downstream tasks \cite{faruqui2016problems,wang2019evaluating,wang2021tsdae}. An illustration is shown in Table~\ref{tab:overfitting-b}. We think there are two reasons behind 1) low testing corpus overlap and 2) mismatch of tested properties.
        
        First, similarity datasets have their data source and are not necessarily close to the corpus of downstream tasks. For example, \citet{baker2014unsupervised} collect word pairs for verbs only while \citet{luong2013better} intentionally test on rare words. Also, for STS datasets, \cite{agirre2012semeval} annotates on sentence pairs from paraphrases, video captions, and machine translations, which has limited overlap on downstream tasks like sentiment classification.
        
        Second, the original goal for the similarity task is to mimic human perceptions. For example, STS datasets are originally proposed as a competition to find the most effective STS systems instead of a gold standard for generic sentence embedding evaluation. Some properties evaluated by similarity tasks are trivial to downstream tasks, and it is more important to test on mutually important ones. As examples in Figure \ref{fig:wspairs}, the similarity tasks inherently require the model to predict sim($p_1$)$>$sim($p_2$) and sim($p_5$)$>$sim($p_4$), which we believe are unnecessary for most downstream applications. Instead, similar pairs are more important than less similar pairs for downstream applications \cite{kekalainen2005binary,reimers2016task}. Therefore, it is enough for good embedding models to focus on gathering similar pairs together while keeping dissimilar ones far away to a certain threshold.
        
    \begin{figure}[t]
        \centering
            \includegraphics[width=0.5\textwidth]{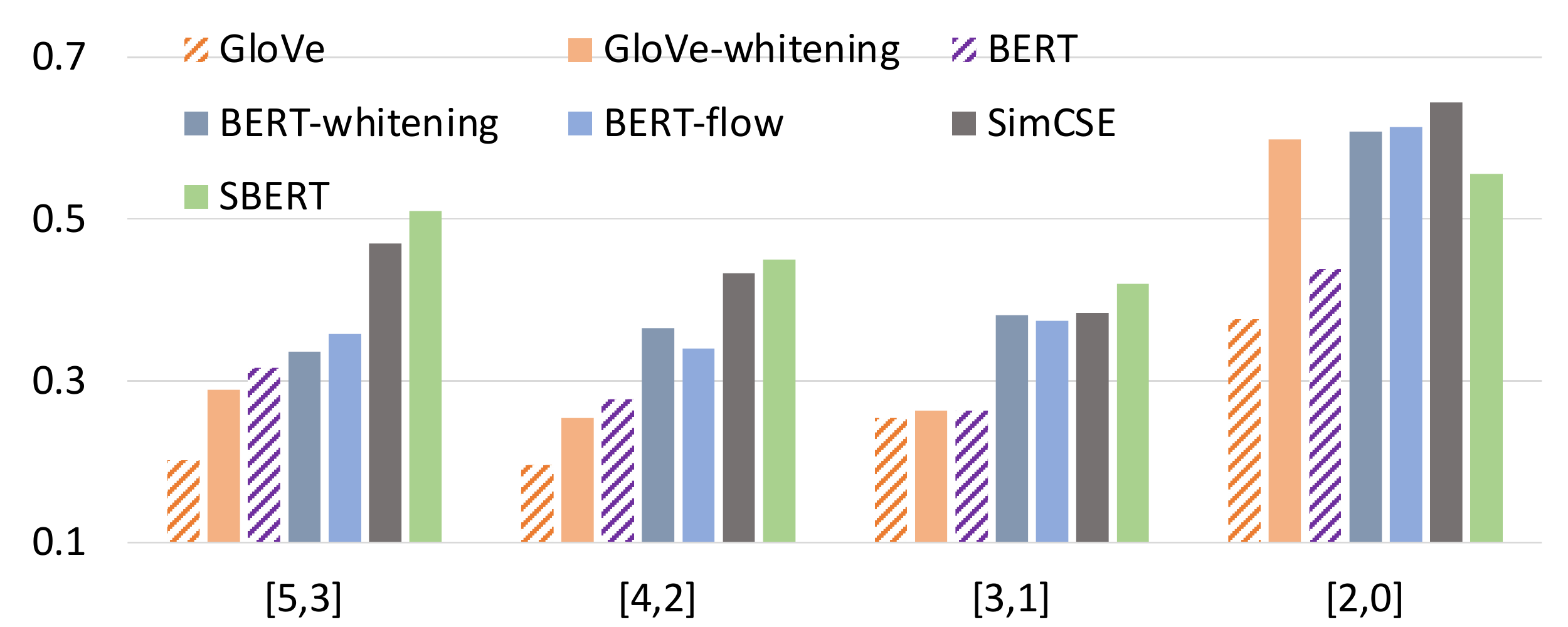}
        \caption{Performance of embedding models on the whole STS-Benchmark dataset w.r.t different similarity levels.}
        \label{fig:stsb_whitening}
    \end{figure}

    \subsection{Overfitting}

        As similarity tasks become one de facto evaluation method for embedding models, recent work tend to overfit the current evaluation paradigm, including the choice of similarity measure and the post-processing step.
        
        \noindent\textbf{Similarity Metrics.} Cosine similarity is the default choice for similarity tasks. However, simply changing the similarity metric to other commonly used ones can lead to contradictory results. 
        
        In Table~\ref{tab:overfitting-a}, we compare recent five BERT-based sentence embedding models including BERT \cite{devlin2018bert}, BERT-whitening \cite{su2021whitening}, BERT-flow \cite{li2020sentence}, SBERT \cite{reimers2019sentence} and SimCSE \cite{gao2021simcse}.\footnote{Experimental details in Appendix \ref{appendix:overfit}.} The results on standard STS-Benchmark testset are reported under both cosine and $l_2$ similarity. As we can see, the performance rank differs under different similarity metrics. This is especially true for BERT-flow and BERT-whitening, which do not even outperform their baseline models when evaluating with $l_2$ metric. Therefore, we can infer that some models overfit to the default cosine metric for similarity tasks.
        
        \noindent\textbf{Whitening Tricks.} A number of studies attempted the post-processing of word embeddings \cite{mu2017all,wang2019post,liu2019unsupervised} and sentence embeddings \cite{arora2017simple,liu2019continual,li2020sentence,su2021whitening}. The shared concept is to obtain a more isotropic embedding space (samples evenly distributed across directions) and can be summarized as a space whitening process. Even though the whitening tricks help a lot with similarity tasks, we found it is usually not applicable to downstream tasks or even hurt the model performance.\footnote{Analysis in Appendix \ref{app:whitening_downstream}.} We think the whitening methods are overfitted to similarity tasks and would like to find the reasons behind.
        
        First, we take the whole STS-Benchmark dataset and create subsets of sentence pairs from certain similarity levels. We test on two baseline sentence embedding models: GloVe, BERT; three whitening tricks: ABTT on GloVe \cite{mu2017all}, BERT-whitening, BERT-flow; two strong sentence embedding models that perform well on both STS and downstream tasks: SBERT, SimCSE. Figure \ref{fig:stsb_whitening} shows the result, and we can see that the whitening-based methods are boosting the baseline performance mainly for less similar pairs (e.g., pairs with a similarity score within [2,0]). In contrast, the models that perform well on downstream tasks show consistent improvement on all subsets with different similarity scores. As discussed in Section \ref{pro:weak_corr}, highly similar pairs are more critical than less similar pairs for downstream tasks. Since the post-processing methods mainly help with less similar pairs, they do not help much on downstream tasks.

    \begin{figure}[t]
        \centering
        \includegraphics[width=0.42\textwidth]{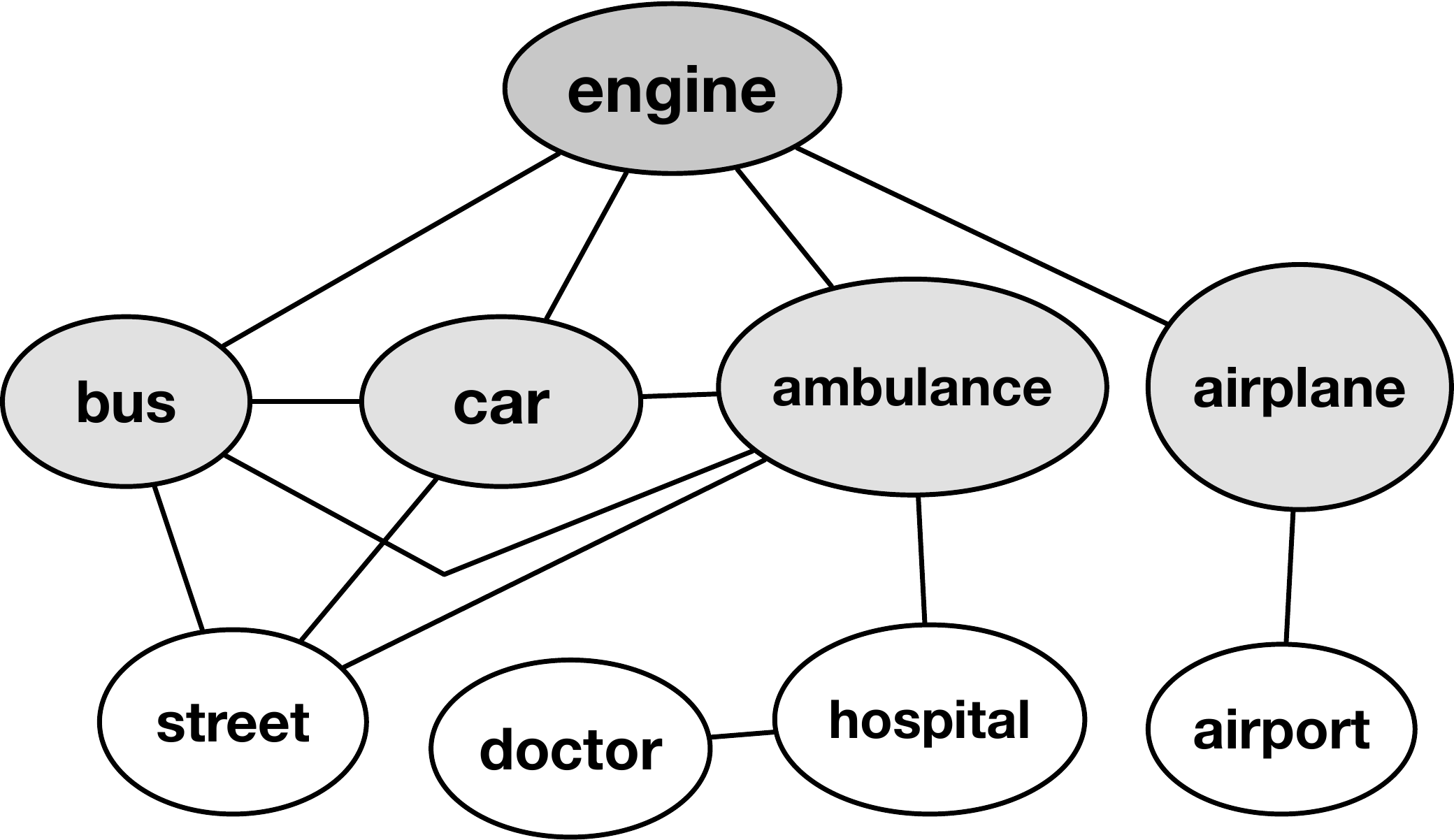}
        \caption{Example of Concept Network in SAT}
        \label{fig:sat}
        \end{figure}

\section{Evaluation by Ranking}

    \begin{table*}[tbh]
        \centering
        \begin{adjustbox}{width=1.0\textwidth,center}
        \begin{tabular}{ l | c | c | c | c  }
        \hline
                           & Type & \# pos pairs & \# background samples & Source    \\
        \hline 
        \multirow{2}{*}{\emph{EvalRank}}     & Word     & 5,514          &  22,207    & Word Similarity Datasets \& Wiki \\ \cline{2-5}
             & Sent         & 6,989     & 24,957     & STS-Benchmark \& STR    \\ 
        \hline
        \end{tabular}
        \end{adjustbox}
        \caption{Statistics of \emph{EvalRank} Datasets}
        \label{tab:dataset}
    \end{table*}

    \subsection{Theory and Motivations}

        In cognitive psychology, Spreading-Activation Theory (SAT) \cite{collins1975spreading,anderson1983spreading} is to explain how concepts store and interact within the human brain. Figure~\ref{fig:sat} shows one example about the concept network. In the network, only highly related concepts are connected. To find the relatedness between concepts like \emph{engine} and \emph{street}, the activation is spreading through mediating concepts like \emph{car} and \emph{ambulance} with decaying factors. Under this theory, the similarity task is measuring the association between any two concepts in the network, which requires complicated long-distance activation propagation. Instead, to test the soundness of the concept network, it is enough to ensure the local connectivity between concepts. Moreover, the long-distance relationships can be inferred thereby with various spreading activation algorithms \cite{cohen1987information}.

        Therefore, we propose \emph{EvalRank} to test only on highly related pairs and make sure they are topologically close in the embedding space. It also alleviates the problems of similarity tasks. First, instead of distinguishing multifaceted relationships, we only focus on highly related pairs, which are intuitive to human annotators. Second, it shows a much stronger correlation with downstream tasks as desired properties are measured. Third, as we treat the embedding space from a local perspective, it is less affected by the whitening methods.

    \subsection{Methodology}

        We frame the evaluation of embeddings as a retrieval task. To this purpose, the dataset of \emph{EvalRank} contains two sets: 1) the positive pair set $P=\{p_1, p_2, ..., p_m\}$ and 2) the background sample set $C = \{c_1, c_2, ..., c_n\}$. Each positive pair $p_i=(c_x,c_y)$ in $P$ consists of two samples in $C$ that are semantically similar.
        
        For each sample ($c_x$) and its positive correspondence ($c_y$), a good embedding model should has their embeddings ($\mathbf{e_x}$, $\mathbf{e_y}$) close in the embedding space. Meantime, the other background samples should locate farther away from the sample $c_x$. Some samples in the background may also be positive samples. We assume it barely happens and is negligible if good datasets are constructed.
        
        Formally, given an embedding model $\phi(\cdot)$, the embeddings for all samples in $C$ are computed as 
        $\{\mathbf{e_1}, \mathbf{e_2}, ...,\mathbf{e_n}\} = \{\phi(c_1), \phi(c_2),...,\phi(c_n)\}$. 
        The $cos$ similarity and $l_2$ similarity between two samples ($c_x, c_y$) are defined as:
        $$S_{cos}(c_x, c_y) = \frac{\mathbf{e}_x^T \mathbf{e}_y}{||\mathbf{e}_x|| \cdot ||\mathbf{e}_y||}$$
        $$S_{l_2}(c_x, c_y) = \frac{1}{1+||\mathbf{e}_x - \mathbf{e}_y ||} $$
        Further, the similarity score is used to sort all background samples in descending order and the performance at each positive pair $p_i$ is measured by the rank of $c_x$'s positive correspondence $c_y$ w.r.t all background samples:
        $$rank_i = rank(S(c_x,c_y), [||_{j=1,j\neq x}^{n} S(c_x,c_j)])$$
        where $||$ refers to the concatenation operation.
        To measure the overall performance of model $\phi(\cdot)$ on all positive pairs in $P$, the mean reciprocal rank (MRR) and Hits@k scores are reported and a higher score indicates a better embedding model: 
        $$MRR = \frac{1}{m}\sum_{i=1}^m \frac{1}{rank_i}$$
        $$Hits@k = \frac{1}{m}\sum_{i=1}^m \mathbbm{1}[rank_i \leq k]$$
        Note that there are two similarity metrics, and we found that $S_{cos}$ shows a better correlation with downstream tasks while $S_{l_2}$ is more robust to whitening methods. We use $S_{cos}$ in the experiments unless otherwise specified.

    \subsection{Dataset Collection}
        
        \noindent\textbf{Word-Level.} We collect the positive pairs from 13 word similarity datasets \cite{wang2019evaluating}. For each dataset, the pairs with the highest 25\% similarity score are gathered as positive pairs. Background word samples contain all words that appear in the similarity datasets. Further, we augment the background word samples using the most frequent 20,000 words from Wikipedia corpus.

        \noindent\textbf{Sentence-Level.} Similarly, the pairs with top 25\% similarity/relatedness score from STS-Benchmark dataset \cite{cer2017semeval} and STR dataset \cite{abdalla2021makes} are collected as positive pairs. All sentences that appear at least once are used as the background sentence samples.
        
        In both cases, if positive pair $(c_x, c_y)$ exists, the reversed pair $(c_y, c_x)$ is also added as positive pairs. Detailed statistics of \emph{EvalRank} datasets are listed in Table~\ref{tab:dataset}.

    \begin{table*}[htb]
        \centering
        \begin{adjustbox}{width=1.0\textwidth,center}
        \begin{tabular}{ c | c | c | c | c | c | c | c | c | c | c | c }
        \hline
        \multicolumn{2}{c|}{}                 & {SCICITE}          & {MR}              & {CR}              & {MPQA}            & {SUBJ}            & {SST2}            & {SST5}            & {TREC}            & {MRPC}            & {SICK-E} \\
        \hline \hline
        \multicolumn{2}{c|}{WS-353-All}   & 62.87              & 43.68             & 40.94             & 37.50             & 15.57             & 41.65             & 45.03             & 34.70             & 8.98              & 57.96 \\ 
        \multicolumn{2}{c|}{WS-353-Rel}       & 66.13              & 47.92             & 45.15             & 41.77             & 11.65             & 47.25             & 48.18             & 26.36             & 20.56             & 61.83 \\ 
        \multicolumn{2}{c|}{WS-353-Sim}       & 67.86              & 45.94             & 43.97             & 38.68             & 17.41             & 44.03             & 50.32             & 34.85             & 10.67             & 56.13 \\ 
        \multicolumn{2}{c|}{RW-STANFORD}      & 75.56              & 74.65             & 55.35             & 66.08             & 46.82             & 81.50             & 68.25             & 45.91             & 13.08             & 43.29 \\ 
        \multicolumn{2}{c|}{MEN-TR-3K}        & 66.91              & 44.15             & 45.37             & 39.14             & 1.70              & 38.51             & 42.11             & 22.82             & 28.63             & \textbf{71.26} \\ 
        \multicolumn{2}{c|}{MTURK-287}        & 68.48              & 65.95             & 48.01             & 52.36             & 31.94             & 71.96             & 58.01             & 29.22             & 7.54              & 36.23 \\ 
        \multicolumn{2}{c|}{MTURK-771}        & 79.93              & 60.87             & 49.45             & 57.92             & 24.04             & 62.75             & 62.03             & 29.14             & 17.44             & 60.23 \\ 
        \multicolumn{2}{c|}{SIMLEX-999}       & 68.20              & 48.02             & 40.90             & 46.43             & 19.03             & 47.30             & 50.95             & 38.14             & 15.32             & 60.26 \\ 
        \multicolumn{2}{c|}{SIMVERB-3500}     & 65.13              & 45.60             & 36.95             & 47.04             & 21.57             & 45.16             & 48.56             & 41.74             & 10.70             & 58.08 \\ 
        \hline
        \multirow{3}{*}{\emph{EvalRank}} & MRR & \underline{89.96}  & \underline{87.91} & \underline{68.23} & 78.03             & 51.35             & \underline{91.54} & \underline{83.36} & \underline{48.15} & 25.70             & 61.34 \\ 
                                    & Hits@1   & 85.91              & 83.69             & 66.93             & \underline{81.43} & \textbf{55.95}    & 89.74             & 79.46             & 43.53             & \underline{28.82} & 53.86 \\
                                    & Hits@3   & \textbf{90.11}     & \textbf{88.82}    & \textbf{69.92}    & \textbf{82.05}    & \underline{54.52} & \textbf{93.32}    & \textbf{84.41}    & \textbf{48.44}    & \textbf{30.87}    & \underline{62.77} \\
        \hline
        \end{tabular}
        \end{adjustbox}
        \caption{Spearman's rank correlation ($\rho \times 100$) between performance scores of word-level intrinsic evaluation and downstream tasks, where the best is marked with \textbf{bold} and second best with \underline{underline}.}
        \label{tab:result_word}
    \end{table*}

    \subsection{Alignment and Uniformity} 
        
        Recently, \citet{wang2020understanding} identifies the alignment and uniformity properties as an explanation to the success of contrastive loss. It shares many similarities with our method and can also shed light on why \emph{EvalRank} works. First, the alignment property requires similar samples to have similar features, which aligns with the objective of \emph{EvalRank}. Second, the uniformity property is measured by the average Gaussian distance between any two samples. In contrast, \emph{EvalRank} focuses on the distance between points from a local perspective and would require the pivot sample to have longer distances to any background samples than its positive candidate. Measuring the distance from a local perspective has unique advantages because the learned embedding space will likely form a manifold and can only approximates euclidean space locally. Therefore, simple similarity metrics like $cos$ or $l_2$ are not suitable to model long-distance relationships.
        
    \subsection{Good Intrinsic Evaluator}
    
        A good intrinsic evaluator can test the properties that semantically similar samples are close in vector space \cite{reimers2019sentence,gao2021simcse} and serve as prompt information to real-world applications. As \emph{EvalRank} directly test on the first property, we design experiments to show the correlation with various downstream tasks as a comparison of intrinsic evaluators. To be comprehensive, we first collect as many embedding models as possible and test them on the intrinsic evaluator and downstream task. The Spearman's rank correlation is computed between the results, and a higher score indicates better correlation with downstream tasks and better intrinsic evaluator.
        
        Meantime, we do not think similarity evaluations should be discarded, even though it fails to correlate well with downstream applications. It has its advantages as aiming to mimic human perception about semantic-related pairs.

\section{Word-Level Experiments}


    \subsection{Experimental Setup}
            
        \noindent\textbf{Word Embedding Models.} We collect 19 word embedding models from GloVe \cite{pennington2014glove}, word2vec \cite{mikolov2013distributed}, fastText \cite{bojanowski2017enriching}, Dict2vec \cite{tissier2017dict2vec} and PSL \cite{wieting2015paraphrase}. Meantime, we apply ABTT \cite{mu2017all} post-processing to all models to double the total number of embedding models. When testing on downstream tasks, the simplest bag-of-words feature is used as sentence representations in order to focus on measuring the quality of word embeddings.

        \noindent\textbf{Word Similarity Tasks.} 9 word similarity datasets are compared as the baseline methods including WS-353-All \cite{finkelstein2001placing}, WS-353-Rel \cite{agirre2009study}, WS-353-Sim \cite{agirre2009study}, RW-STANFORD \cite{luong2013better}, MEN-TR-3K \cite{bruni2014multimodal}, MTURK-287 \cite{radinsky2011word}, MTURK-771 \cite{halawi2012large}, SIMLEX-999 \cite{hill2015simlex}, SIMVERB-3500 \cite{gerz2016simverb}. The word similarity datasets with less than 200 pairs are not selected to avoid evaluation occasionality. 
        Cosine similarity and Spearman's rank correlation are deployed for all similarity tasks.

        \noindent\textbf{Downstream Tasks.} 
        SentEval \cite{conneau2018senteval} is a popular toolkit in evaluating sentence embeddings. We use 9 downstream tasks from SentEval including MR \cite{pang2005seeing}, CR \cite{hu2004mining}, MPQA \cite{wiebe2005annotating}, SUBJ \cite{pang2004sentimental}, SST2 \cite{socher2013recursive}, SST5 \cite{socher2013recursive}, TREC \cite{li2002learning}, MRPC \cite{dolan2004unsupervised}, SICK-E \cite{marelli2014sick}. 
        Previous work spot that SentEval tasks are biased towards sentiment analysis \cite{wang2018glue}. Therefore, we add one extra domain-specific classification task SCICITE \cite{cohan2019structural} which assigns intent labels (background information, method, result comparison) to sentences collected from scientific papers that cite other papers.
        For all tasks, a logistic regression classifier is used with cross-validation to predict the class labels.    
        
        \begin{table}[t]
            \centering
            \begin{adjustbox}{width=0.5\textwidth,center}
                \begin{tabular}{ l | c | c | c }
                \hline
                                            & SCICITE       & MR        & SST2      \\ \hline \hline
                \emph{EvalRank}             & 89.96         & 87.91     & 91.54     \\ \hline
                    w/o\ \ \  wiki vocabs   & 88.55         & 83.99     & 88.26     \\ 
                    w/ \quad WN synonym        & 90.56         & 86.56     & 91.12      \\ 
                    w/ \quad $l_2$ metric   & 77.47         & 78.34     & 81.51     \\ 
                \hline
                \end{tabular}
            \end{adjustbox}
            \caption{Ablation study on variants of \emph{EvalRank}. Spearman's rank correlation ($\rho \times 100$) between MRR scores and downstream task scores are reported.}
            \label{tab:er-wl-ab}
        \end{table}
        
    \subsection{Results and Analysis}

        Table~\ref{tab:result_word} shows the word-level results. In short, \emph{EvalRank} outperforms all word similarity datasets with a clear margin. For evaluation metrics, we can see that Hits@3 score shows a higher correlation than MRR and Hits@1 scores. However, the gap between the evaluation metrics is not big, which makes them all good measures. Among all 10 downstream tasks, \emph{EvalRank} shows a strong correlation ($\rho$>0.6) with 7 tasks and a very strong correlation ($\rho$>0.8) with 5 tasks. While, among all word similarity datasets, only one dataset (RW-STANFORD) shows a strong correlation with one downstream task (SST2).

        For word similarity datasets, RW-STANFORD dataset shows the best correlation with downstream tasks. It confirms the finding in \citet{wang2019evaluating} that this dataset contains more high-quality and low-frequency word pairs.

        \noindent\textbf{Ablation Study.} We experiment with several variants of our \emph{EvalRank} method and the result is shown in Table~\ref{tab:er-wl-ab}. First, if we do not augment the background word samples with the most frequent 20,000 words from the Wikipedia corpus, it leads to certain performance downgrading. Without sufficient background samples, positive pairs are not challenging enough to test each model's capability. Second, we tried to add more positive samples (e.g. 5k samples) using synonym relations from WordNet (WN) database \cite{miller1998wordnet}. However, no obvious improvement is witnessed because the synonym pairs in WN contain too many noisy pairs. Last, for similarity measures, we notice that $cos$ similarity is consistently better than $l_2$ similarity while both outperform word similarity baselines.

        \noindent\textbf{Benchmarking Results.} 
            In Table~\ref{tab:benchmarking-word}, we compared four popular word embedding models, including GloVe, word2vec, fastText, and Dict2vec, where fastText achieves the best performance.

        \begin{table}[t]
            \centering
            \begin{subtable}[b]{0.43\textwidth}
                \centering
                \begin{adjustbox}{width=1.0\textwidth,center}
                \begin{tabular}{l c c c}
                    \hline
                    \emph{EvalRank} & MRR & Hits@1 & Hits@3 \\
                    \hline \hline
                    GloVe & 13.15 & 4.66 & 15.72 \\
                    word2vec & 12.88 & 4.57 & 14.35 \\
                    fastText & \textbf{17.22} & \textbf{5.77} & \textbf{19.99} \\
                    Dict2vec & 12.71 & 4.03 & 13.04 \\
                    \hline
                \end{tabular}
                \end{adjustbox}
                \caption{Word-Level}
                \label{tab:benchmarking-word}
            \end{subtable}
            \\
            \begin{subtable}[b]{0.5\textwidth}
                \centering
                \begin{adjustbox}{width=1.0\textwidth,center}
                \begin{tabular}{l c c c}
                    \hline
                    \emph{EvalRank} & MRR & Hits@1 & Hits@3 \\
                    \hline\hline
                    GloVe & 61.00 & 44.94 & 74.66 \\
                    InferSentv1 & 60.72 & 41.92 & 77.21 \\
                    InferSentv2 & 63.89 & 45.59 & 80.47 \\
                    BERT-first-last-avg & 68.01 & 51.70 & 81.91 \\
                    BERT-whitening & 66.58 & 46.54 & 84.22 \\
                    SBERT & 64.12 & 47.07 & 79.05 \\
                    SimCSE & \textbf{69.50} & \textbf{52.34} & \textbf{84.43} \\
                    \hline
                \end{tabular}
                \end{adjustbox}
                \caption{Sentence-Level}
                \label{tab:benchmarking-sent}
            \end{subtable}
        \caption{Benchmarking results on \emph{EvalRank}. Performance is reported as \% ($\times 100$).}
        \label{tab:benchmarking}
        \end{table}

        \begin{table*}[t]
            \centering
            \begin{adjustbox}{width=1.0\textwidth,center}
            \begin{tabular}{ c | c | c | c | c | c | c | c | c | c  }
            \hline
            \multicolumn{2}{c|}{}                  & {SCICITE}         & {MR}              & {CR}              & {MPQA}            & {SUBJ}                & {SST2}            & {SST5}            & {TREC}     \\
            \hline \hline
            \multicolumn{2}{c|}{STS12}             & 32.96             & 38.62             & 44.77             & 31.52             &  21.76                &  33.79            & 35.68             & 30.79  \\ 
            \multicolumn{2}{c|}{STS13}             & 22.04             & 32.62             & 41.23             & 12.39             &  7.64                 &  26.45            & 22.98             & 12.16  \\ 
            \multicolumn{2}{c|}{STS14}             & 25.91             & 34.77             & 41.89             & 19.23             &  10.13                &  29.20            & 26.82             & 17.70  \\ 
            \multicolumn{2}{c|}{STS15}             & 31.84             & 40.64             & 48.11             & 25.12             &  16.48                &  35.50            & 33.30             & 24.70  \\ 
            \multicolumn{2}{c|}{STS16}             & 29.56             & 40.14             & 51.66             & 14.35             &  16.53                &  33.61            & 29.44             & 21.43  \\ 
            \multicolumn{2}{c|}{STS-Benchmark}     & 32.99             & 46.03             & 52.78             & 21.09             &  26.47                &  40.41            & 36.75             & 34.64  \\ 
            \multicolumn{2}{c|}{SICK-Relatedness}  & 40.38             & 38.51             & 50.68             & 29.87             &  18.87                &  34.54            & 36.73             & 25.25  \\ 
            \multicolumn{2}{c|}{STR}               & -14.48            & -8.38             & -7.79             & -29.57            & -23.91                &  -16.33           & -22.77            & -14.30 \\
            \hline
            \multirow{3}{*}{\emph{EvalRank}} & MRR  & \underline{65.95} & 83.43             & \underline{87.08} & \underline{43.93} &  \underline{72.72}    & \underline{80.97} & \underline{74.16} & \underline{76.74} \\
            & Hits@1                                & \textbf{69.01}    & \textbf{85.39}    & \textbf{89.36}    & \textbf{45.81}    &  \textbf{74.93}       & \textbf{82.65}    & \textbf{76.65}    & \textbf{78.72}   \\
            & Hits@3                                & 63.35             & \underline{83.92} & 85.43             & 41.24             &  70.98                & 80.36             & 72.05             & 74.70  \\
                
            \hline
            \end{tabular}
            \end{adjustbox}
            \caption{Spearman's rank correlation ($\rho \times 100$) between performance scores of sentence-level intrinsic evaluation and downstream tasks, where the best is marked with \textbf{bold} and second best with \underline{underline}.}
            \label{tab:sent-results}
            \end{table*}

\section{Sentence-Level Experiments}

    \subsection{Experimental Setup}
            
        \noindent\textbf{Sentence Embedding Models.} We collect 67 embedding models, where 38 of them are built upon word embeddings with bag-of-words features and 29 of them are neural-network-based models. For neural-network-based models, we collect variants from InferSent \cite{conneau2017supervised}, BERT \cite{devlin2018bert}, RoBERTa \cite{liu2019roberta}, BERT-flow \cite{li2020sentence}, BERT-whitening \cite{su2021whitening}, SBERT \cite{reimers2019sentence} and SimCSE \cite{gao2021simcse}.

        \noindent\textbf{Sentence Similarity Tasks.} We evaluate on 7 standard semantic textual similarity datasets including STS12$\sim$16 \cite{agirre2012semeval,agirre2013sem,agirre2014semeval,agirre2015semeval,agirre2016semeval}, STS-Benchmark \cite{cer2017semeval} and SICK-Relatedness \cite{marelli2014sick}. 
        Recently, \citet{abdalla2021makes} questioned the labeling process of STS datasets and released a new semantic textual relatedness (STR) dataset, which is also included in our experiments.

        \noindent\textbf{Downstream Tasks.} We use 7 classification tasks from SentEval evaluation toolkit, including MR, CR, MPQA, SUBJ, SST2, SST5, TREC, as well as the domain-specific classification task SCICITE. We exclude the MRPC and SICK-E because they are highly similar with STS tasks \cite{conneau2018senteval}.

    \subsection{Results and Analysis}

        Table~\ref{tab:sent-results} shows the sentence-level results. \emph{EvalRank} outperform all sentence similarity datasets with a clear margin. For evaluation metric, Hits@1 shows a higher correlation comparing with MRR and Hits@3. Among all 7 downstream tasks, \emph{EvalRank} shows strong correlation ($\rho > 0.6$) with 6 tasks.

        \begin{figure}[t]
            \centering
            \includegraphics[width=0.5\textwidth]{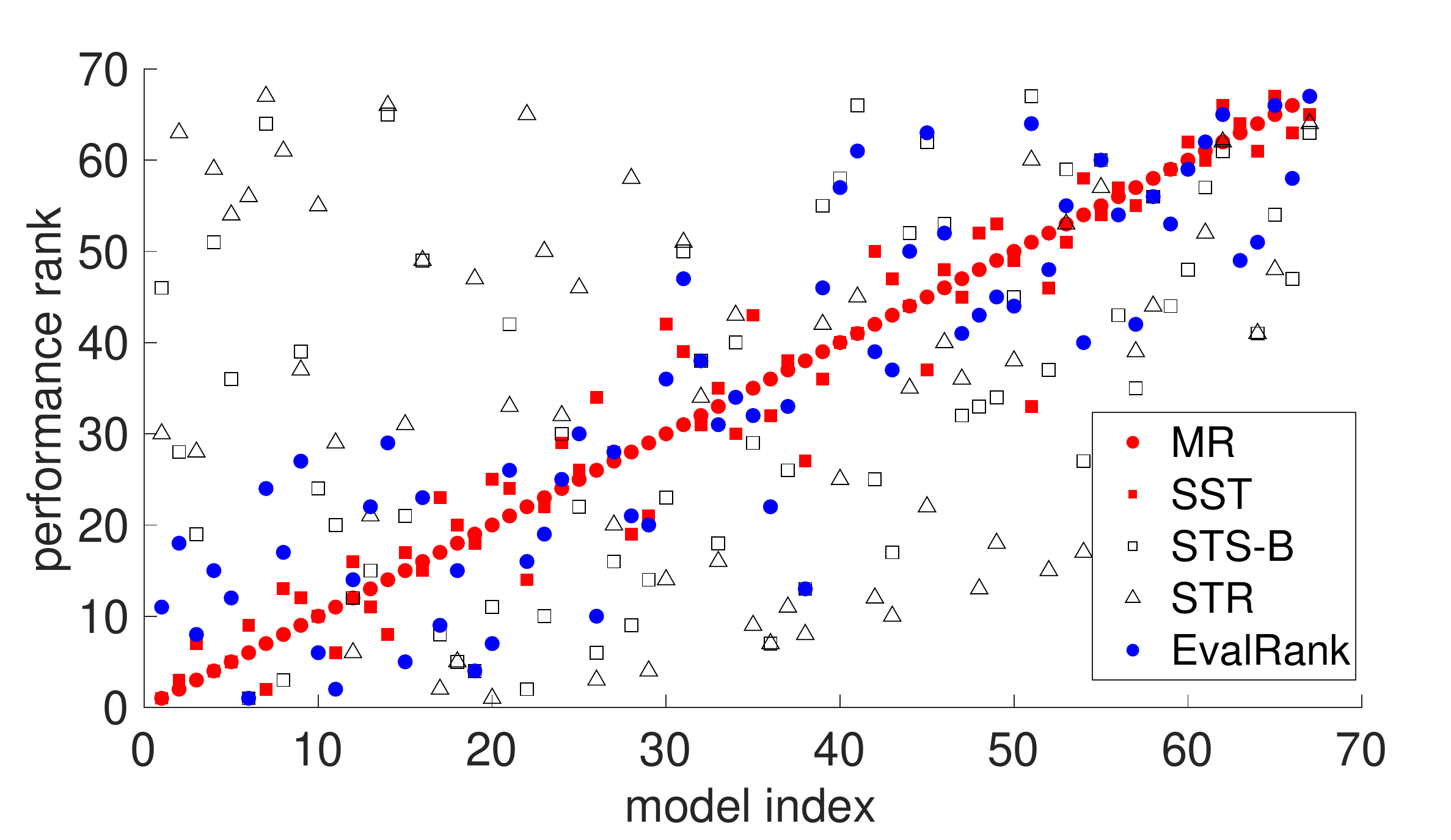}
            \caption{Visualization of models' performance rank on 2 downstream tasks and 3 intrinsic evaluation methods.}
            \label{fig:sent-rank}
            \end{figure}

        \begin{table}[t]
            \centering
            \begin{adjustbox}{width=0.5\textwidth,center}
            \begin{tabular}{ c | c | c | c | c }
            \hline
            \multicolumn{2}{c|}{}                                          & SCICITE       &  MR           & SST2 \\ \hline \hline
            \multirow{3}{*}{\shortstack[c]{\emph{EvalRank}\\(STS-B +\\STR)}} & MRR   & 65.95         & 83.43         & 80.97 \\ 
            & Hits@1                                                        & 69.01         & 85.39         & 82.65 \\
            & Hits@3                                                        & 63.35         & 83.92         & 80.36 \\ \hline
            
            \multirow{3}{*}{\shortstack[c]{\emph{EvalRank}\\(STS-B)}} & MRR   & 63.05         & 75.85         & 72.87 \\ 
            & Hits@1                                                        & 66.22         & 77.94         & 75.20 \\
            & Hits@3                                                        & 61.23         & 75.49         & 72.92 \\ \hline
            
            \multirow{3}{*}{\shortstack[c]{\emph{EvalRank}\\(STR)}} & MRR   & 63.51         & 83.28         & 80.20 \\ 
            & Hits@1                                                        & 66.59         & 84.53         & 82.14 \\
            & Hits@3                                                        & 60.68         & 82.55         & 79.42 \\ \hline
            
            \hline
        \end{tabular}
        \end{adjustbox}
        \caption{Performance under different data sources. Spearman's rank correlation ($\rho \times 100$) is reported.}
        \label{tab:var_sent_er}
        \end{table}

        For sentence similarity datasets, no one clearly outperforms others. 
        Additionally, we found that STR dataset shows the worst correlation with downstream tasks. Even though STR adopts a better data annotation schema than STS datasets, it still follows the previous standard evaluation paradigm and is exposed to the same problems.
        It further verifies our discussion about problems with sentence similarity evaluation.
            
        \noindent\textbf{Correlation Visualization.} 
        Figure~\ref{fig:sent-rank} shows the performance rank of 67 sentence embedding models on five tasks, including 2 downstream tasks (MR, SST2) and 3 intrinsic evaluations (STS-B, STR, \emph{EvalRank}). The models' performance rank on the MR task is used as the pivot.

        As MR and SST2 datasets are both related to sentiment analysis, they correlate well with each other. Among the three intrinsic evaluation tasks, \emph{EvalRank} shows a higher correlation with downstream tasks as the blue dots roughly follow the trend of red dots. In contrast, the dots of STS-B and STR are dispersed in different regions. This shows that the performance of STS-B and STR is not a good indicator of the performance on downstream tasks.

        \noindent\textbf{Ablation Study.}
        In Table~\ref{tab:var_sent_er}, we show the performance of \emph{EvalRank} with different data sources. By combining the positive pairs collected from both STS-B and STR datasets, \emph{EvalRank} leads to the best performance. Interestingly, according to our results, even though STR evaluation does not correlate well with downstream tasks, the positive pairs collected from STR have better quality than STS-B. It also confirms the argument that STR improves the dataset collection process \cite{abdalla2021makes}.

        \noindent\textbf{Benchmarking Results.} Table~\ref{tab:benchmarking-sent} benchmarked seven popular sentence embedding models. As the widely accepted SOTA model, SimCSE outperforms others with a clear margin.

\section{Conclusion}

    In this work, we first discuss the problems with current word and sentence similarity evaluations and proposed \emph{EvalRank}, an effective intrinsic evaluation method for word and sentence embedding models. It shows a higher correlation with downstream tasks. We believe that our evaluation method can have a broader impact in developing future embedding evaluation methods, including but not limited to its multilingual and task-specific extensions.



\section*{Acknowledgement}
    
    This research is supported by the Agency for Science, Technology and Research (A*STAR) under its AME Programmatic Funding Scheme (Project No. A18A2b0046) and Science and Engineering Research Council, Agency of Science, Technology and Research (A*STAR), Singapore, through the National Robotics Program under Human-Robot Interaction Phase 1 (Grant No. 192 25 00054).

\bibliography{anthology,custom}
\bibliographystyle{acl_natbib}

\appendix


\section{Embedding Models}
    
    A good intrinsic evaluator should be a good indicator for downstream tasks. We want to compute the correlation between the results from intrinsic evaluators and downstream tasks to measure the quality of intrinsic evaluators. For this purpose, we collect as many models as possible and finally involved 38 word embedding models and 67 sentence embedding models in our experiments. We give a detailed introduction to the collected embedding models in this section.
    
    A complete set of selected word embedding models is shown in Table~\ref{tab:we-models}. We collect pre-trained word embeddings with different dimensions and training corpus from GloVe, word2vec, fastText, Dict2vec, and PSL. ABTT \cite{mu2017all} post-processing is further applied to each model to double the total number of word embedding models.

    A complete set of selected sentence embedding models is shown in Table~\ref{tab:se-models}. Besides the models obtained using bag-of-words features from word embeddings, we also include popular neural-network-based models including InferSent, BERT, RoBERTa, SBERT, BERT-whitening, BERT-flow, and SimCSE. Different variants of these models are considered in order to be more comprehensive.

\section{More Experimental Details}
    \label{appendix:overfit}
    
    In Section~\ref{sec:problems}, we conduct several experiments to certify our judgments, and we would like to elaborate on the detailed experiment settings here.
    
    In Table~\ref{tab:overfitting-a}, the performance rank of five BERT-based sentence embedding models are shown under both $cos$ and $l_2$ distance measure. Detailed model settings are shown below:
        
        \begin{itemize}
            
            \item \textbf{SBERT}: BERT-base model trained on Natural Language Inference data with mean token embeddings. \\
            \url{https://huggingface.co/sentence-transformers/bert-base-nli-mean-tokens} 
            
            \item \textbf{SimCSE}: Unsupervised SimCSE trained upon BERT-based-uncased. \\
            \url{https://huggingface.co/princeton-nlp/unsup-simcse-bert-base-uncased}
            
            \item \textbf{BERT}: BERT-based uncased model \\
            \url{https://huggingface.co/bert-base-uncased}
            
            \item \textbf{BERT-flow}: We use the BERT-base-uncased and average the word representations from the first and last layers as the sentence representation. The Gaussian mapping is trained on the target corpus, which is the STS-B testset in our case.
            
            \item \textbf{BERT-whitening}: Similar to BERT-flow, the averaging of word representation from first and last layers are used as sentence representations, and the BERT-base-uncased model is used. The whitening objective is computed using the target corpus.
            
        \end{itemize}
        
    In Table~\ref{tab:overfitting-b}, 6 models are selected, and their performance on one similarity task: STS-B and two downstream tasks: MR and SST2 are reported. The setting of the models follows the experiments in Table~\ref{tab:overfitting-a}. For GloVe and InferSent, the following settings are used:
    
        \begin{itemize}
            
            \item \textbf{GloVe}: 300-dimensional vector trained on Common Crawl corpus (840B tokens).
            
            \item \textbf{InferSent}: Version 1 of InferSent is used where the GloVe model is served as input.
            
        \end{itemize}
    
    Figure~\ref{fig:stsb_whitening} shows a detailed analysis on different similarity levels. For the experiment, we first collect all sentence pairs from STS-B dataset. Then, we split the pairs into four subsets based on their similarity levels ([5,3],[4,2],[3,1],[2,0]). Further, we randomly sampled 3,000 samples for each subset as the final dataset splits to keep the number of samples even.

\section{More Discussions}

    \subsection{Effect of Whitening on Downstream Tasks}
    \label{app:whitening_downstream}
    
        A lot whitening methods been proposed targeting on improving the quality of word embeddings \cite{mu2017all,wang2019post,liu2019unsupervised} and sentence embeddings \cite{arora2017simple,liu2019continual,li2020sentence,su2021whitening}. However, in previous studies, the whitening methods are only proven to be effective with similarity tasks. The performance comparison on downstream tasks is either missing or limited.
        
        Therefore, we conduct extensive experiments on two popular post-processing methods. For word embedding, the ABTT \cite{mu2017all} post-processing technique is examined. For sentence embedding, the Principal Component Removal \cite{arora2017simple} method is applied for word-embedding-based models and BERT-whitening \cite{su2021whitening} or BERT-flow \cite{li2020sentence} is applied to BERT-based models. \citet{arora2017simple} propose a weighting schema and post-processing step for sentence embeddings. Here, we solely test the effectiveness of the post-processing step.
        
        Table~\ref{tab:whiten-downstream} shows the performance comparison between the original model and the post-processed model. From both word-level and sentence-level experiments, we conclude that the post-processing methods play no obvious role or even hurt the performance in downstream tasks. In contrast, the results on similarity tasks improve a lot.

    \subsection{Alignment and Uniformity}
        
        \citet{wang2020understanding} discussed alignment and uniformity property as an explanation to the success of contrastive learning. \emph{EvalRank} can be viewed as a variant of these two measures and focus more on the local perspective. Therefore, the success of \emph{EvalRank} also can be explained under the same umbrella. Meantime, measuring from a local perspective is more suitable for word and sentence embedding models because they are likely to form a manifold and can only approximate euclidean space locally.

        \noindent\textbf{Alignment}: In \citet{wang2020understanding}, the alignment loss is defined with the average distance between positive samples:
        $$L_{align}(f;\alpha)=\mathbb{E}_{(x,y)\sim p_{pos}}[||f(x)-f(y)||_2^\alpha]$$
        It measures the total distance between positive pairs, and the smaller, the better. The alignment measure does not consider the local properties of the embedding space. In contrast, \emph{EvalRank} requires the positive pairs to be close in the embedding space while considering the density of the local embedding regions. If the density of embedding space around positive pairs is high, \emph{EvalRank} method requires the embeddings of positive pairs to be more tightly closed. If the density of embedding space around positive pairs is low, \emph{EvalRank} has a looser distance requirement for the positive pairs. 

        \noindent\textbf{Uniformity}: In \citet{wang2020understanding}, the uniformity loss is designed as the logarithm of the average pairwise Gaussian potential:
        $$L_{uniform}(f;t)=\log\ \ \mathbb{E}_{(x,y\tilde p_{data})}[e^{-t||f(x)-f(y)||_2^2}]$$
        Intuitive, the uniformity loss asks features to be far away from each other. In contrast, \emph{EvalRank} score focus on a local perspective. It requires the negative samples to have larger embedding distances than positive samples concerning the pivot sample. For the negative samples that are far away from the pivot sample in the embedding space, they are less likely to be confusing with positive samples and, therefore, not considered as important.

    \subsection{Correlation Results without Post-Processing Models}
        
        In previous experiments, we select as many models as possible in order to be more comprehensive. However, the side effect is that a reasonable portion of the models is built with post-processing techniques. It may lead to some concern that our selected embedding models might be biased on post-processed models. Therefore, we re-do the experiments on sentence embedding evaluations without considering post-processed models.

        We filter out all models related to post-processing techniques, and as a result, 34 sentence embedding models are kept. We further conduct correlation analysis between the performance on intrinsic evaluation methods and downstream tasks. 
        
        The result is shown in Table~\ref{tab:sent-results-less-models}. As we can see, \emph{EvalRank} still outperforms sentence similarity tasks in 7 of the tasks. And we can witness a higher correlation between \emph{EvalRank} and the downstream tasks comparing with the results in Table~\ref{tab:sent-results}. \emph{EvalRank} shows strong correlation ($\rho > 0.6$) on all 8 tasks and very strong correlation ($\rho > 0.8$) on 7 of the tasks. The result again proves the effectiveness of \emph{EvalRank}.

    \begin{table*}[htb]
        \centering
        \begin{adjustbox}{width=1.0\textwidth,center}
        \begin{tabular}{| l | l | l | l |}
        \hline
        Model \# & Model Name & Details & Post-process \\\hline \hline
        1 / 2  & GloVe \cite{pennington2014glove} & glove.6B.50d & no / yes \\\hline
        3 / 4  & GloVe & glove.6B.100d & no / yes\\\hline
        5 / 6  & GloVe & glove.6B.200d & no / yes \\\hline
        7 / 8  & GloVe & glove.6B.300d & no / yes \\\hline
        9 / 10  & GloVe & glove.42B.300d & no / yes \\\hline
        11 / 12 & GloVe & glove.840B.300d & no / yes \\\hline
        13 / 14 & GloVe & glove.twitter.27B.25d & no / yes \\\hline
        15 / 16 & GloVe & glove.twitter.27B.50d & no / yes \\\hline
        17 / 18 & GloVe & glove.twitter.27B.100d & no / yes \\\hline
        19 / 20 & GloVe & glove.twitter.27B.200d & no / yes \\\hline
        21 / 22 & word2vec \cite{mikolov2013distributed} & GoogleNews-vectors-negative300 & no / yes \\\hline
        23 / 24 & fastText \cite{bojanowski2017enriching} & crawl-300d-2M & no / yes \\\hline
        25 / 26 & fastText & crawl-300d-2M-subword & no / yes \\\hline
        27 / 28 & fastText & wiki-news-300d-1M & no / yes \\\hline
        29 / 30 & fastText & wiki-news-300d-1M-subword & no / yes \\\hline
        31 / 32 & Dict2vec \cite{tissier2017dict2vec} & dict2vec-100d & no / yes \\\hline
        33 / 34 & Dict2vec & dict2vec-200d & no / yes \\\hline
        35 / 36 & Dict2vec & dict2vec-300d & no / yes \\\hline
        37 / 38 & PSL \cite{wieting2015paraphrase} & paragram\_300\_sl999 & no / yes \\\hline
        \end{tabular}
        \end{adjustbox}
        \caption{Word embedding models used in our evaluation. We use ABTT as the post-processing method \cite{mu2017all}.}
        \label{tab:we-models}
    \end{table*}

    \begin{table*}[htb]
        \centering
        \begin{adjustbox}{width=1.0\textwidth,center}
        \begin{tabular}{ | l | l | l | l | l | }
        \hline
        Model \# & Model Type & Model Name & Details & Post-Process \\\hline \hline
        1 / 2  & we-bow & GloVe & glove.6B.50d & no / yes \\\hline
        3 / 4  & we-bow & GloVe & glove.6B.100d & no / yes \\\hline
        5 / 6  & we-bow & GloVe & glove.6B.200d & no / yes \\\hline
        7 / 8  & we-bow & GloVe & glove.6B.300d & no / yes \\\hline
        9 / 10  & we-bow & GloVe & glove.42B.300d & no / yes \\\hline
        11 / 12  & we-bow & GloVe & glove.840B.300d & no / yes \\\hline
        13 / 14  & we-bow & GloVe & glove.twitter.27B.25d & no / yes \\\hline
        15 / 16  & we-bow & GloVe & glove.twitter.27B.50d & no / yes \\\hline
        17 / 18  & we-bow & GloVe & glove.twitter.27B.100d & no / yes \\\hline
        19 / 20  & we-bow & GloVe & glove.twitter.27B.200d & no / yes \\\hline
        21 / 22  & we-bow & word2vec & GoogleNews-vectors-negative300 & no / yes \\\hline
        23 / 24  & we-bow & fasttext & crawl-300d-2M & no / yes \\\hline
        25 / 26  & we-bow & fasttext & crawl-300d-2M-subword & no / yes \\\hline
        27 / 28  & we-bow & fasttext & wiki-news-300d-1M & no / yes \\\hline
        29 / 30  & we-bow & fasttext & wiki-news-300d-1M-subword & no / yes \\\hline
        31 / 32  & we-bow & dict2vec & dict2vec-100d & no / yes \\\hline
        33 / 34  & we-bow & dict2vec & dict2vec-200d & no / yes \\\hline
        35 / 36  & we-bow & dict2vec & dict2vec-300d & no / yes \\\hline
        37 / 38  & we-bow & PSL & paragram\_300\_sl999 & no / yes \\\hline
        39 / 40  & neural net & InferSent & v\_1 / v\_2 & no \\\hline
        41 / 42  & neural net & BERT & bert-base-uncased \& cls & no / whitening \\\hline
        43 / 44  & neural net & BERT & bert-base-uncased \& last-avg & no / whitening \\\hline
        45 / 46  & neural net & BERT & bert-base-uncased \& first-last-avg & no / whitening \\\hline
        47 / 48  & neural net & BERT & bert-large-uncased \& cls & no / whitening \\\hline
        49 / 50  & neural net & BERT & bert-large-uncased \& last-avg & no / whitening \\\hline
        51 / 52  & neural net & BERT & bert-large-uncased \& first-last-avg & no / whitening \\\hline
        53 / 54  & neural net & RoBERTa & roberta-base \& last-avg & no / whitening \\\hline
        55 / 56  & neural net & RoBERTa & roberta-base \& first-last-avg & no / whitening \\\hline
        57 / 58  & neural net & RoBERTa & roberta-large \& last-avg & no / whitening \\\hline
        59 / 60  & neural net & RoBERTa & roberta-large \& first-last-avg & no / whitening \\\hline
        61  & neural net & BERT-flow & bert-base-uncased \& cls & N/A \\\hline
        62  & neural net & BERT-flow & bert-base-uncased \& last-avg & N/A \\\hline
        63  & neural net & BERT-flow & bert-base-uncased \& first-last-avg & N/A \\\hline
        64 / 65 & neural net & SBERT & sbert-base-nli-mean-tokens & no / whitening \\\hline
        66  & neural net & SimCSE & unsup-simcse-bert-base-uncased & no \\\hline
        67  & neural net & SimCSE & sup-simcse-bert-base-uncased & no \\\hline
        \end{tabular}
        \end{adjustbox}
        \caption{Sentence embedding models used in our evaluation. For word-embedding-based models, the bag-of-words feature is used, and the principal component removal algorithm is used as the post-processing of sentence embeddings \cite{arora2017simple}. For post-processing for BERT-based model, the BERT-whitening model is applied \cite{su2021whitening}}
        \label{tab:se-models}
    \end{table*}

        \begin{table*}[htb]
        \centering
        \begin{adjustbox}{width=0.6\textwidth,center}
        \begin{tabular}{| c || c || c |}
        \hline
                        & Word-Level                                & Sentence-Level                        \\\hline 
        Post-Process    & No v.s. Yes                               & No v.s. Yes                           \\ \hline \hline
        SCICITE         & 45.0\% \textcolor{red}{<} 55.0\%          & 67.6\% \textcolor{green}{>} 32.4\%    \\ \hline
        MR              & 42.1\% \textcolor{red}{<} 57.9\%          & 63.6\% \textcolor{green}{>} 36.4\%    \\ \hline
        CR              & 26.3\% \textcolor{red}{<} 73.7\%          & 54.5\% \textcolor{green}{>} 45.5\%    \\ \hline
        MPQA            & 94.7\% \textcolor{green}{>} 5.3\%         & 97.0\% \textcolor{green}{>} 3.0\%     \\ \hline
        SUBJ            & 78.9\% \textcolor{green}{>} 21.1\%        & 87.9\% \textcolor{green}{>} 12.1\%    \\ \hline
        SST2            & 80.0\% \textcolor{green}{>} 20.0\%        & 88.2\% \textcolor{green}{>} 11.8\%    \\ \hline
        SST5            & 80.0\% \textcolor{green}{>} 20.0\%        & 88.2\% \textcolor{green}{>} 11.8\%    \\ \hline
        TREC            & 89.5\% \textcolor{green}{>} 10.5\%        & 81.8\% \textcolor{green}{>} 18.2\%    \\ \hline
        MRPC            & 50.0\% = 50.0\%                           & 64.7\% \textcolor{green}{>} 35.3\%    \\ \hline
        SICK-E          & 15.0\% \textcolor{red}{<} 85.0\%          & 51.5\% \textcolor{green}{>} 48.5\%    \\ \hline \hline
        WS-353-All      & 21.1\% \textcolor{red}{<} 78.9\%          & NA                                    \\ \hline
        WS-353-Rel      & 26.3\% \textcolor{red}{<} 73.7\%          & NA                                    \\ \hline
        WS-353-Sim      & 21.1\% \textcolor{red}{<} 78.9\%          & NA                                    \\ \hline
        RW-STANFORD     & 47.4\% \textcolor{red}{<} 52.6\%          & NA                                    \\ \hline
        MEN-TR-3K       & 15.8\% \textcolor{red}{<} 84.2\%          & NA                                    \\ \hline
        MTURK-287       & 26.3\% \textcolor{red}{<} 73.7\%          & NA                                    \\ \hline
        MTURK-771       & 21.1\% \textcolor{red}{<} 78.9\%          & NA                                    \\ \hline
        SIMLEX-999      & 21.1\% \textcolor{red}{<} 78.9\%          & NA                                    \\ \hline
        SIMVERB-3500    & 36.8\% \textcolor{red}{<} 63.2\%          & NA                                    \\ \hline\hline
        STS12           & NA                                        & 3.0\% \textcolor{red}{<} 97.0\%       \\\hline
        STS13           & NA                                        & 0.0\% \textcolor{red}{<} 100.0\%      \\\hline
        STS14           & NA                                        & 0.0\% \textcolor{red}{<} 100.0\%      \\\hline
        STS15           & NA                                        & 0.0\% \textcolor{red}{<} 100.0\%      \\\hline
        STS16           & NA                                        & 3.0\% \textcolor{red}{<} 97.0\%       \\\hline
        STS-Benchmark   & NA                                        & 0.0\% \textcolor{red}{<} 100.0\%      \\\hline
        SICK-Relatedness& NA                                        & 15.2\% \textcolor{red}{<} 84.8\%      \\\hline
        STR             & NA                                        & 6.0\% \textcolor{red}{<} 94.0\%       \\\hline 
        \end{tabular}
        \end{adjustbox}
        \caption{Performance of models with and without post-processing step.}
        \label{tab:whiten-downstream}
        \end{table*}

        \begin{table*}[htb]
            \centering
            \begin{adjustbox}{width=0.9\textwidth,center}
            \begin{tabular}{ c | c | c | c | c | c | c | c | c | c  }
            \hline
            \multicolumn{2}{c|}{}                  & {SCICITE}          & {MR}              & {CR}              & {MPQA}            & {SUBJ}                & {SST2}            & {SST5}            & {TREC}     \\
            \hline \hline
            \multicolumn{2}{c|}{STS12}             & 35.45              & 39.07             & 39.65             & 63.36             & 28.60                 & 30.87             & 42.92             & 42.58      \\ 
            \multicolumn{2}{c|}{STS13}             & 42.51              & 42.88             & 46.71             & \textbf{72.68}    & 32.44                 & 34.10             & 47.70             & 42.15      \\ 
            \multicolumn{2}{c|}{STS14}             & 37.99              & 38.05             & 41.73             & 68.05             & 27.50                 & 30.18             & 43.38             & 43.16      \\ 
            \multicolumn{2}{c|}{STS15}             & 44.19              & 46.07             & 47.41             & 68.78             & 35.08                 & 38.11             & 51.14             & 50.13      \\ 
            \multicolumn{2}{c|}{STS16}             & 63.62              & 64.30             & 66.33             & \underline{71.81} & 56.87                 & 57.09             & 68.38             & 66.03      \\ 
            \multicolumn{2}{c|}{STS-Benchmark}     & 47.10              & 48.82             & 51.05             & 62.93             & 38.26                 & 40.98             & 53.76             & 54.22      \\ 
            \multicolumn{2}{c|}{SICK-Relatedness}  & 49.98              & 51.65             & 54.90             & 68.44             & 41.18                 & 42.57             & 57.04             & 54.82      \\ 
            \multicolumn{2}{c|}{STR}               & -2.57              & -1.53             & -7.81             & 27.70             & -1.47                 & -5.42             & -3.12             & 9.37      \\
            \hline
            \multirow{3}{*}{\emph{EvalRank}} & MRR & 83.37              & \underline{85.40} & \underline{85.45} & 66.38             & 81.29                 & \underline{82.62} & \underline{85.77} & 85.69      \\
            & Hits@1                               & \underline{83.69}  & \textbf{86.15}    & \textbf{85.82}    & 65.80             & \textbf{82.28}        & \textbf{83.21}    & \textbf{85.78}    & \textbf{86.67}      \\
            & Hits@3                               & \textbf{83.92}     & 85.19             & 84.97             & 66.38             & \underline{81.37}     & 82.36             & 85.74             & \underline{86.51}      \\
            \hline
            \end{tabular}
            \end{adjustbox}
            \caption{Spearman's rank correlation ($\rho \times 100$) between performance scores of sentence-level intrinsic evaluation and downstream tasks, where the best is marked with \textbf{bold} and second best with \underline{underline}. The models with post-processing are filtered out, resulting in a total number of 34 sentence embedding models.}
            \label{tab:sent-results-less-models}
        \end{table*}


\end{document}